\documentclass[letterpaper, 10 pt, conference]{ieeeconf}  %

\IEEEoverridecommandlockouts                              %

\usepackage{amsmath} %
\usepackage{amssymb}  %

\usepackage[noadjust]{cite}

\makeatletter
\let\NAT@parse\undefined
\makeatother

\usepackage{graphicx}   %
\usepackage[dvipsnames]{xcolor}
\definecolor{hgreen}{RGB}{28, 164, 98}

\usepackage{kotex}
\usepackage{lipsum}
\usepackage{pifont}%
\usepackage[normalem]{ulem}
\usepackage{mathtools}

\usepackage{array}
\newcolumntype{P}[1]{>{\centering\arraybackslash}p{#1}}
\newcolumntype{M}[1]{>{\centering\arraybackslash}m{#1}}
\newcolumntype{L}[1]{>{\hspace{0.5em}\raggedright\arraybackslash}m{#1}}
\newcolumntype{R}[1]{>{\raggedleft\arraybackslash}m{#1}}

\usepackage{xspace}

\makeatletter
\DeclareRobustCommand\onedot{\futurelet\@let@token\@onedot}
\def\@onedot{\ifx\@let@token.\else.\null\fi\xspace}

\makeatother

\usepackage[colorlinks=true, linkcolor=blue, citecolor=blue, urlcolor=blue, pagebackref=true]{hyperref}
\definecolor[named]{ACMDarkBlue}{cmyk}{1,0.58,0,0.21}
\definecolor{darkgreen}{RGB}{25,200,25}
\hypersetup{colorlinks,
  linkcolor=ACMDarkBlue,  %
  citecolor=ACMDarkBlue,  %
  urlcolor=ACMDarkBlue,   %
  filecolor=ACMDarkBlue
}

\usepackage{cleveref}
\Crefformat{figure}{Fig.~#2#1#3}
\Crefformat{equation}{(#2#1#3)}
\Crefformat{table}{Tab.~#2#1#3}

\usepackage{caption}

\DeclareCaptionFormat{tablewithgap}{#1#2\vspace{3pt}#3}
\usepackage{tabularx,booktabs}
\newcolumntype{Y}{>{\centering\arraybackslash}X}
\usepackage{colortbl}
\usepackage{multirow}
\usepackage{soul}
\setul{0.4ex}{0.4pt}   %
\usepackage{wrapfig}
\newcommand{\drule}{\specialrule{0.2pt}{1pt}{1pt}%
            \specialrule{0.2pt}{0pt}{\belowrulesep}%
            }

\usepackage{footmisc}
\newcommand{\dagfootnote}[1]{%
  \textsuperscript{\dag}%
  \begingroup
    \let\savedfootnoterule\footnoterule
    \renewcommand{\thefootnote}{}%
    \renewcommand{\footnoterule}{%
      \vspace{1ex}\hrule width 0.4\textwidth\vspace{1ex}%
    }%
    \footnotetext{\dag\ #1}%
    \addtocounter{footnote}{-1}%
    \let\footnoterule\savedfootnoterule
  \endgroup
}

\usepackage{algorithm,algpseudocode}
\usepackage[algo2e]{algorithm2e} 
\definecolor{commentcolor}{RGB}{110,154,155}
\definecolor{defcolor}{RGB}{225,81,145}

\usepackage{titletoc}

\makeatletter
\let\labelindent\relax
\makeatother

\usepackage{enumitem}
\usepackage{flushend}

\newcommand{\std}[1]{{\scriptsize$\pm$#1}}

\definecolor{ourc}{RGB}{96,171,112}
\newcommand{\ourmodel}{SeeR-VLA\xspace}

\newcommand{\g}{\cellcolor{gray!12}}

\title{\LARGE \bf
{\color{ourc}See} like a {\color{ourc}R}obot: Robot-Centric Pointmaps for {\color{ourc}VLA} Models
}

\author{
Byungkun Lee$^{1,*,\dagger}$, Dongyoon Hwang$^{1,*}$,
Dongjin Kim$^{1}$, Hojoon Lee$^{2}$,\\
Hyunseung Kim$^{1,3}$, Jaegul Choo$^{1}$, and Minho Park$^{1}$\\[0.4em]
{\normalsize $^{1}$KAIST AI \quad $^{2}$Holiday Robotics \quad $^{3}$KRAFTON AI}%
\thanks{$^{*}$These authors contributed equally.}%
\thanks{$^{\dagger}$Corresponding author.}%
}

\usepackage{needspace}

\begin{document}

\maketitle
\thispagestyle{empty}
\pagestyle{empty}

\begin{abstract}
Vision-language-action (VLA) models require 3D spatial reasoning, yet RGB observations encode robot-object geometry only implicitly.
Lifting depth with camera intrinsics makes this geometry explicit as dense, image-aligned pointmaps, but their camera-frame coordinates depend on camera placement.
We propose \ourmodel, which transforms pointmaps into a robot-centric frame with an end-effector origin and robot-base-aligned axes.
An encoder initialized from pretrained RGB weights extracts pointmap features, which are added to corresponding RGB tokens without increasing the token count.
Across 24 RoboCasa tasks and four real-world tasks, \ourmodel improves average success over RGB-only $\pi_{0.5}$ by 6.4 and 32.5 percentage points, respectively.
It also exceeds the strongest evaluated 3D-augmented baseline, PointVLA, by 3.5 and 23.7 percentage points, respectively.
Beyond these gains, our ablations clarify how coordinate choices affect VLA performance, showing that end-effector centering is most effective with robot-base-aligned axes.
The benefits grow as training viewpoints diversify, highlighting the importance of using robot-frame pointmaps when learning from diverse camera configurations. The project page is available \href{https://davian-robotics.github.io/pointmap/}{here}.
\end{abstract}

\section{Introduction}
\label{sec:intro}

\begin{figure*}
\centering
\includegraphics[width=\linewidth]{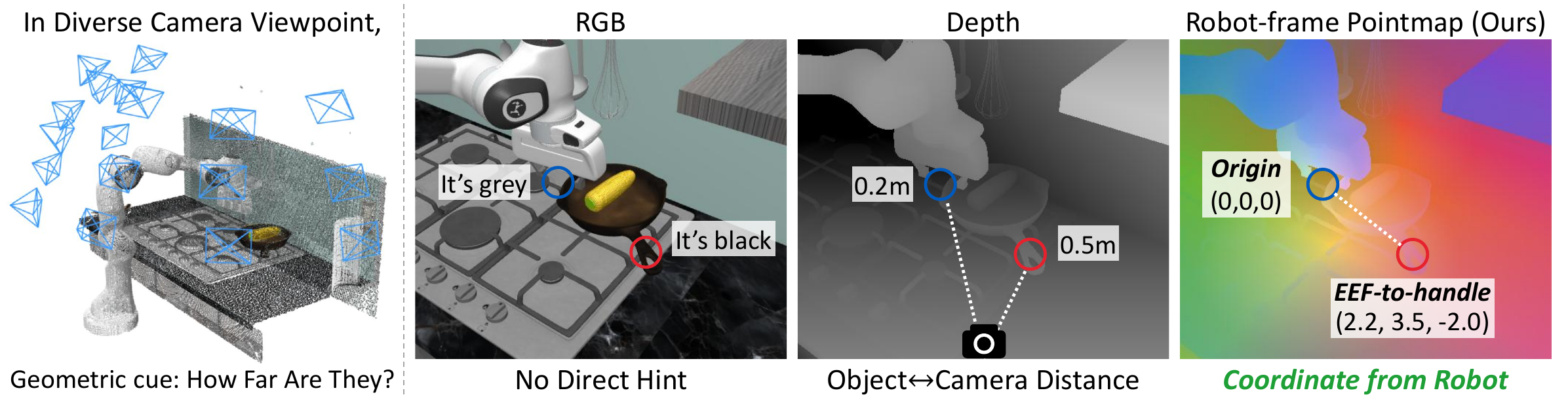}
\vspace{0.05cm}
\caption{\textbf{Depth versus robot-centric pointmaps.}
Depth measures camera-relative distance, whereas our pointmap stores pixel-aligned 3D coordinates relative to the end effector with robot-base-aligned axes, making the represented interaction geometry invariant to training-time camera viewpoint.}
\label{fig:motivation}
\vspace{-0.3cm}
\end{figure*}

Vision-language-action (VLA) models learn robotic manipulation policies from large-scale datasets, predicting actions from visual observations and language instructions~\cite{zitkovich2023rt2, bjorck2025gr00t, black2025pi05}.
Accurate action prediction requires reasoning about metric 3D geometry, such as the target object's position relative to the gripper during grasping.
However, as shown in \Cref{fig:motivation}, RGB observations encode this geometry only implicitly, requiring policies to infer object poses and their spatial relationships to the gripper from robot data with limited direct geometric supervision.
This gap has motivated growing interest in equipping VLA models with explicit 3D information~\cite{4dvla, 3dcavla, goyal2024rvt2, pointvla, geovla, fan2026any3d, hwang20263dhamster}.

Depth is widely used to provide explicit 3D information for manipulation~\cite{3dcavla,4dvla,hwang20263dhamster,ke20243ddiffuseractor,gkanatsios20253dfa,goyal2024rvt2}, but a scalar depth value alone does not specify a pixel's 3D position.
Recovering this position requires back-projecting the pixel using its depth and camera intrinsics.
The resulting 3D coordinates are commonly processed as point clouds, which explicitly represent scene geometry but are often downsampled into sparse point sets to reduce computational cost.
This sampling can discard fine geometric details needed for precise manipulation.
Pointmaps instead preserve the dense output of depth back-projection by storing each pixel's 3D coordinates in an image-aligned $H \times W \times 3$ grid.
Constructed directly from depth and camera intrinsics, this representation retains dense geometry and allows standard image encoders to combine it with visual features at corresponding locations.
These properties have motivated increasing use of pointmaps in 3D vision~\cite{dust3r,moge,moge2,vggt,cut3r,pi3} and robot policy learning~\cite{pointmappolicy,remapdp,yan2026flexpi}.

However, pointmaps express 3D positions in the camera frame, so the same scene point has different coordinates when observed from different camera placements.
This variation is particularly relevant to VLAs pretrained on large-scale robot datasets collected with diverse camera configurations~\cite{o2024openxembodiment, bu2025agibot_iros, khazatsky2024droid}.
A common solution is to express pointmaps in the robot base frame, making their coordinates independent of camera placement~\cite{pointmappolicy}.
For manipulation, however, the target's position relative to the end effector is particularly important~\cite{feng2026demystifying}.
Centering pointmaps at the end effector directly captures this relationship by describing grasp targets and articulated parts relative to a common interaction origin~\cite{framemining}.
This representation preserves the coordinates of the same interaction geometry across changes in camera placement or interaction location within the workspace, providing a consistent geometric reference for learning manipulation.

To this end, we propose \ourmodel, which augments pretrained VLAs with robot-centric pointmaps that combine explicit interaction geometry with image-aligned visual features.
We construct pointmaps from depth and camera calibration, placing the origin at the end effector while keeping the coordinate axes aligned with the robot base.
Each point thus describes a displacement from the gripper without changing the axis directions as the gripper rotates.
To incorporate this geometry into the VLA, we process pointmaps with a separate encoder initialized from the pretrained RGB encoder.
The shared image grid allows us to add pointmap features to RGB features at corresponding spatial locations, pairing visual appearance with the position of the observed surface relative to the end effector.
These fused features replace the original RGB tokens, providing explicit 3D information without increasing the visual token count or changing the policy's action prediction objective.

We evaluate \ourmodel on 24 RoboCasa tasks and four real-world tasks collected with diverse training camera viewpoints.
\ourmodel outperforms PointVLA~\cite{pointvla}, the strongest evaluated 3D-augmented VLA baseline, by 3.5 percentage points on RoboCasa and 23.7 percentage points on real-world tasks.
Controlled ablations also show that explicitly constructing pointmaps is more effective than supplying depth and camera cues for the policy to learn the geometric conversion, and that retaining the image grid outperforms point-cloud encoding.
The best results are obtained by combining an end-effector origin, robot-base-aligned axes, and additive RGB-pointmap fusion.
Importantly, the advantage over RGB-only policies grows as training camera viewpoints become more diverse, supporting the value of a consistent geometric representation.
Masking and action-sensitivity analyses further reveal that the policy selectively relies on target-object and robot coordinates rather than irrelevant background geometry.

We make the following contributions.
\begin{itemize}[leftmargin=*]
    \item We introduce \ourmodel, which augments VLAs with robot-centric pointmaps through RGB-pointmap fusion for manipulation across diverse camera viewpoints.
    \item We identify an effective pointmap design through controlled ablations, combining an end-effector origin, robot-base-aligned axes, and additive RGB-pointmap fusion on the image grid.
    \item We show that robot-centric pointmaps improve robustness to training-time viewpoint variation, with increasing gains over RGB-only policies as viewpoints diversify.
\end{itemize}

\section{Related Work}
\label{sec:related}

\subsubsection{3D Geometric Observation for Robotics}

Robotic manipulation policies use diverse 3D representations to capture geometry beyond RGB appearance.
Depth provides dense, image-aligned cues~\cite{3dcavla,4dvla,hwang20263dhamster}, but recovering explicit 3D coordinates requires back-projection using camera intrinsics.
Point-cloud methods process the resulting geometry~\cite{yuan2023m2t2,ze2024dp3,ke20243ddiffuseractor,gkanatsios20253dfa,fan2026any3d,pointact}, often subsampling it for efficiency and potentially discarding fine geometric details. 
Other approaches use voxel grids~\cite{peract}, 3D feature fields~\cite{gervet2023act3d}, virtual views~\cite{goyal2024rvt2}, or unified multi-view representations~\cite{robouniview}.
Pointmaps instead organize back-projected 3D coordinates in a dense $H\times W\times3$ image grid, preserving RGB-pixel correspondence and allowing standard image encoders to process the geometry.
These properties have motivated their use in 3D vision~\cite{dust3r,mast3r,moge,moge2,vggt,cut3r,fast3r,pi3} and robot policy learning~\cite{pointmappolicy,remapdp,yan2026flexpi}.

Coordinate frames also determine how this geometry is expressed to the policy.
PointMapPolicy~\cite{pointmappolicy} represents dense pointmaps in a shared world frame, while FP3~\cite{fp3} uses robot-base-frame point clouds, aligning metric coordinates across camera views.
However, geometrically similar interactions can still be represented by different coordinates when they occur at different workspace locations.
Frame Mining~\cite{framemining} systematically studies coordinate-frame choices for pointcloud-based manipulation and shows that frames beyond the robot-base frame, such as end-effector and target-centered frames, can improve policy learning.
Related work uses object-centric trajectories~\cite{chopsticks}, affordance-relative states and actions~\cite{orientedaffordance}, or canonicalized point-cloud inputs~\cite{canonicalpolicy}.
Adapt3R~\cite{adapt3r} transforms semantic point clouds into the end-effector coordinate frame.
We study coordinate-frame design for dense pointmaps in pretrained VLAs, separately varying the origin and axis orientation to distinguish end-effector centering from rotation of the axes with the gripper.

\subsubsection{Injecting 3D Geometry into VLAs}
Beyond selecting a geometric representation and reference frame, incorporating geometry into pretrained VLAs requires deciding how it interacts with existing features. Camera-aware methods provide calibration-derived cues~\cite{posevla,jiang2026knowyourcamera} or express actions in the camera frame~\cite{ocvla}, whereas scene-geometry-aware methods enrich visual representations with depth-derived spatial embeddings~\cite{spatialvla,4dvla}. Other approaches introduce dedicated geometric encoders or experts~\cite{geovla,pointvla,3dcavla,depthvla,fan2026any3d,pointact,hwang20263dhamster}, render geometry into image inputs~\cite{bridgevla,ogvla}, or incorporate future 3D scene generation~\cite{3dvla}. Geometric foundation models offer another route through feature injection~\cite{evo0} or training-time representation alignment~\cite{spatialforcing}. Despite these advances, the integration of image-aligned pointmaps into pretrained VLAs remains underexplored. We investigate this integration by initializing the pointmap encoder with pretrained RGB-encoder weights and comparing fusion strategies, without requiring point-cloud-specific architectures or additional rendering or scene-generation stages. 

\begin{figure*}
\centering
\includegraphics[width=\linewidth]{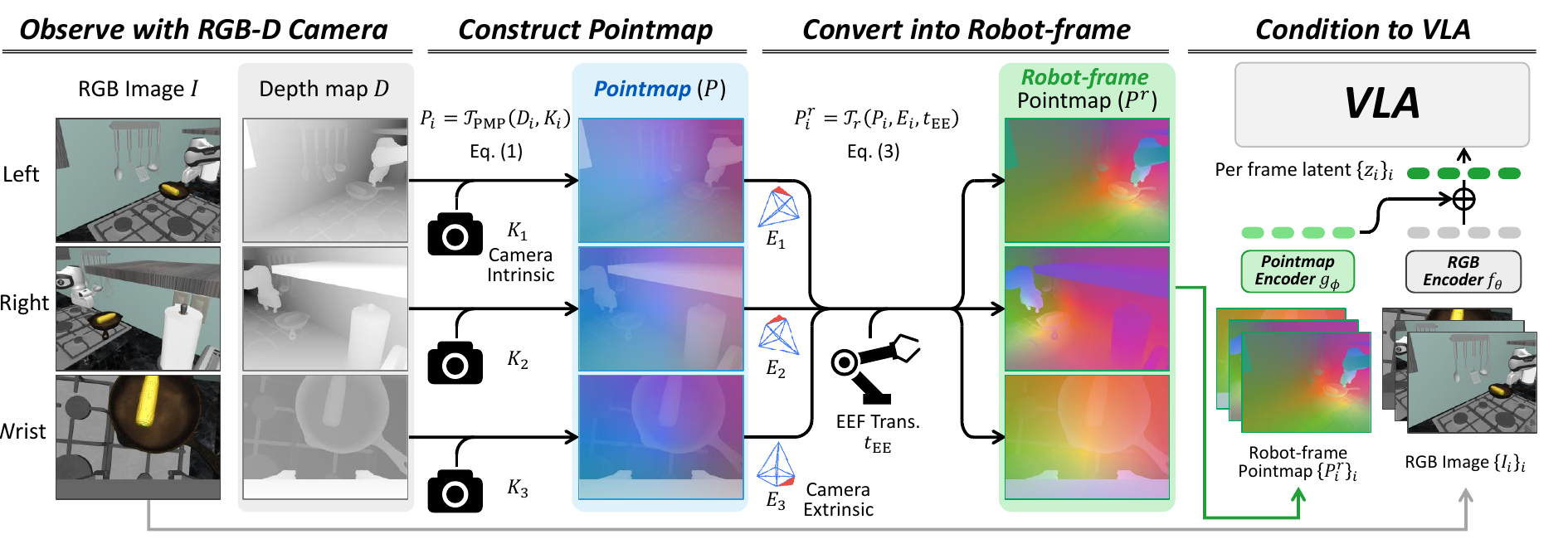}
\caption{\textbf{Overview of the \ourmodel framework.}
We lift depth into camera-frame pointmaps using intrinsics, then construct robot-frame pointmaps by applying extrinsics and shifting the origin to the end effector while retaining robot-base-aligned axes. Next, the RGB encoder $f_\theta$ and pointmap encoder $g_\phi$ extract corresponding features.
Finally, we add these features to form $z_i$, replacing the VLA's RGB tokens without increasing the token count.}
\label{fig:overview}
\vspace{-0.4cm}
\end{figure*}

\section{Method}
\label{sec:method}

To bridge the coordinate-frame mismatch between camera-centric observations and robot-centric actions, \ourmodel converts multi-view RGB-D observations into explicit 3D pointmaps expressed in the robot coordinate frame, as illustrated in \Cref{fig:overview}.
We first introduce the problem setup and notation used throughout the method (\Cref{sec:3-notation}).
We then construct a pointmap for each view from its depth observation and camera intrinsics (\Cref{sec:3-construct-pointmap}), and transform the resulting pointmaps into a common robot coordinate frame (\Cref{sec:3-robot-centric-pointmap}).
Finally, we condition the VLA on these robot-centric pointmaps to provide explicit 3D geometric information aligned with the coordinate frame in which robot actions are executed (\Cref{sec:3-fusion}).

\subsection{Problem Setup}
\label{sec:3-notation}

We consider a multi-view robotic manipulation setting with $N$ RGB-D cameras. At each timestep, the robot observes RGB images $\{I_i\}_{i=1}^{N}$ and their corresponding depth maps $\{D_i\}_{i=1}^{N}$.
For each camera $i$, we assume known intrinsics $K_i$ and extrinsics $E_i \coloneqq T^{r \leftarrow c_i}$, where
$T^{r \leftarrow c_i}$ denotes the transformation from the $i$-th camera coordinate frame to the robot coordinate frame.
The robot proprioceptive state is denoted by $S$, from which the end-effector pose $(R_{\mathrm{EE}}, t_{\mathrm{EE}})$ is obtained via forward kinematics.
We build on the VLA policy $\pi_{\text{0.5}}$~\cite{black2025pi05}, augmenting its visual input with robot-frame pointmaps.
The language instruction, robot proprioceptive state, and flow-matching training objective~\cite{liu2022flow} otherwise follow the original VLA setup~\cite{black2025pi05}.

\subsection{Pointmap Construction}
\label{sec:3-construct-pointmap}

Given the depth map $D_i$ and camera intrinsics $K_i$ of the $i$-th view, we construct a camera-frame pointmap $P_i$ as
\begin{equation}
P_i = \mathcal{T}_{\mathrm{PMP}}(D_i, K_i)
\in \mathbb{R}^{H \times W \times 3},
\label{eq:construct_pmp}
\end{equation}
where $\mathcal{T}_{\mathrm{PMP}}$ denotes the pixel-to-3D lifting operation.
Specifically, for each pixel $(u,v)$,
\begin{equation}
P_i(u,v)
=
D_i(u,v) K_i^{-1}
\begin{bmatrix}
u \\
v \\
1
\end{bmatrix}.
\label{eq:cameraframe_pmp}
\end{equation}
Each element of $P_i$ therefore stores the 3D coordinate of the corresponding pixel in the $i$-th camera coordinate frame.
The resulting pointmap preserves the $H \times W$ spatial grid of the RGB image, providing a pixel-aligned geometric representation.

\subsection{Robot-Centric Pointmap}
\label{sec:3-robot-centric-pointmap}

The pointmap $P_i$ is expressed in the local coordinate frame of the $i$-th camera.
To obtain a view-consistent geometric representation aligned with the robot, we transform each pointmap into a robot-centric coordinate frame.
Using the camera-to-robot extrinsic $E_i$ and the current end-effector position $t_{\mathrm{EE}}$, we define the robot-centric pointmap as
\begin{equation}
P_i^{r}
=
\mathcal{T}_{r}(P_i, E_i, t_{\mathrm{EE}}),
\label{eq:robotcentric_pmp}
\end{equation}
where $\mathcal{T}_{r}$ first transforms the camera-frame coordinates into the robot base frame and then shifts the origin to the current end effector.

Let $R_i$ and $t_i$ denote the rotation and translation components of $E_i$, respectively.
For each point in the pointmap, the transformation is given by
\begin{equation}
P_i^{r}(u,v)
=
R_i P_i(u,v) + t_i - t_{\mathrm{EE}}.
\label{eq:robotframe_pmp}
\end{equation}

Since all pointmaps are expressed in the same robot-centric coordinate system, a physical scene point has the same coordinate across training-time camera viewpoints.
Moreover, centering the representation at the current end effector expresses each scene point by its displacement from the gripper rather than by its absolute location in the workspace.
Consequently, interactions requiring similar end-effector motions induce similar local geometry even when they occur at different workspace locations.
We ablate the effect of this robot-centric representation in \Cref{sec:origin_ablation}.

\begin{figure*}
\centering
\includegraphics[width=\linewidth]{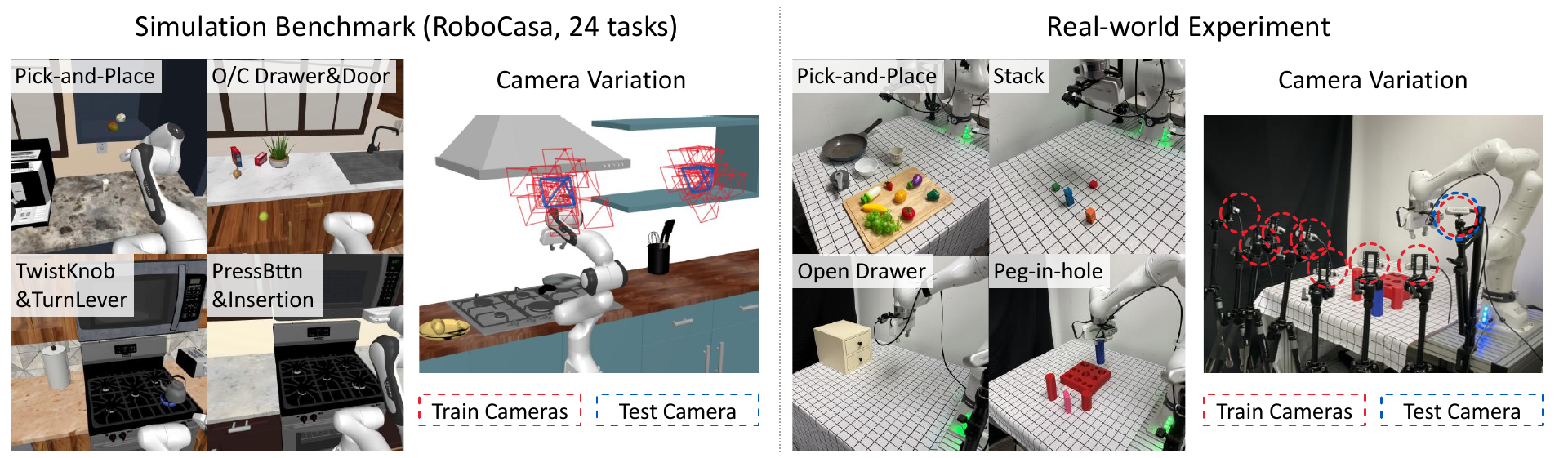}
\caption{\textbf{Simulation and real-world experimental setups.}
Left, we evaluate on 24 RoboCasa tasks with camera placements randomized across training demonstrations (red) and fixed to a reference viewpoint during evaluation (blue).
Right, our real-world tasks comprise pick-and-place, block stacking, drawer opening, and peg-in-hole insertion.
Demonstrations are collected from multiple camera placements (red), and evaluation uses the rightmost placement (blue).}
\label{fig:setup}
\end{figure*}

\subsection{Pointmap-RGB Fusion}
\label{sec:3-fusion}

Since the pointmap preserves the $H \times W$ spatial grid of the RGB image, it can be processed using the same encoder architecture as the RGB observation.
We introduce a separate pointmap encoder $g_{\phi}$ that shares the architecture of the VLA's RGB encoder $f_{\theta}$ and is initialized with its pretrained weights.
The two encoders produce tokens of the same shape, allowing us to fuse the geometric and visual features through element-wise addition:
\begin{equation}
z_i
=
f_{\theta}(I_i)
+
g_{\phi}(P_i^{r})
\in
\mathbb{R}^{N_{\mathrm{tok}} \times d},
\label{eq:add}
\end{equation}
where $N_{\mathrm{tok}}$ and $d$ denote the number and dimension of the visual tokens, respectively.
The fused tokens $z_i$ replace the original RGB tokens as the visual input to the VLA.
Because the fusion is performed through element-wise addition, the number of visual tokens remains unchanged, allowing the pretrained VLA backbone to incorporate robot-centric 3D geometry without increasing its input sequence length.

\section{Experiments}
\label{sec:main}
Our experiments address the following questions.
\begin{itemize}[leftmargin=*, nosep]
    \item \textit{RQ1:} How much do robot-centric pointmaps improve manipulation performance over RGB-only policies and existing 3D-augmented VLAs? (\Cref{sec:main-exp})
    \item \textit{RQ2:} How do different 3D representations, pointmap-RGB fusion methods, and training-time viewpoint settings affect VLA performance? (\Cref{sec:ablation})
    \item \textit{RQ3:} How should the pointmap coordinate frame be defined to best support VLAs? (\Cref{sec:origin_ablation})
   \item \textit{RQ4:} Which pointmap regions does \ourmodel rely on for action prediction? (\Cref{sec:analysis-sec})
\end{itemize}

\begin{table*}[t]
\centering
\footnotesize
\setlength{\tabcolsep}{3pt}
\captionsetup{labelsep=period, textfont=small, labelfont=small}
\caption{
Comparison with 3D-augmented VLAs on 24 RoboCasa and four real-world tasks.
}
\label{tab:main_robocasa}
\vspace{-0.15cm}
\begin{tabular}{l cccc|c | cccc|c}
\toprule
& \multicolumn{5}{c|}{RoboCasa~\cite{robocasa2024}} & \multicolumn{5}{c}{Real-world} \\
\cmidrule(lr){2-6} \cmidrule(lr){7-11}
Method & Pick\&Place & Open/Close & Knob/Lever & Button/Insert & Avg. & Pick \& place & Stack & Open dr. & Peg in hole & Avg. \\
\drule
FP3~\cite{fp3} & 24.3\,\std{2.3} & 76.1\,\std{1.3} & 33.2\,\std{1.8} & 42.4\,\std{3.1} & 42.9\,\std{0.7} & 25.0 & 15.0 & \ul{80.0} & \ul{55.0} & 43.8 \\
\midrule
$\pi_{0.5}$~\cite{black2025pi05} & 45.2\,\std{0.8} & 81.6\,\std{2.1} & 41.6\,\std{2.4} & 52.7\,\std{1.4} & 55.1\,\std{0.3} & 50.0 & 15.0 & 65.0 & 20.0 & 37.5 \\
\rowcolor{gray!12}
\;\;+ KYC~\cite{jiang2026knowyourcamera} & 47.0\,\std{5.2} & 83.1\,\std{3.2} & 43.5\,\std{5.1} & 49.6\,\std{5.7} & 55.8\,\std{3.9} & --- & --- & --- & --- & --- \\
\rowcolor{gray!12}
\;\;+ OC-VLA~\cite{ocvla} & \ul{52.8}\,\std{2.4} & 80.8\,\std{0.8} & \textbf{46.4}\,\std{2.8} & 48.0\,\std{2.9} & 57.4\,\std{1.1} & \ul{60.0} & 25.0 & 75.0 & 20.0 & 45.0 \\
\rowcolor{yellow!12}
\;\;+ GeoVLA~\cite{geovla} & 50.4\,\std{0.8} & 81.2\,\std{1.0} & 42.1\,\std{4.8} & 55.1\,\std{2.2} & 57.4\,\std{1.4} & --- & --- & --- & --- & --- \\
\rowcolor{yellow!12}
\;\;+ PointVLA~\cite{pointvla} & 47.7\,\std{1.8} & \ul{86.6}\,\std{0.2} & 42.5\,\std{2.8} & \ul{55.7}\,\std{1.0} & \ul{58.0}\,\std{0.7} & 45.0 & \ul{30.0} & 75.0 & 35.0 & \ul{46.3} \\
\rowcolor{ourc!12}
\;\;+ \ourmodel (Ours) & \textbf{52.9}\,\std{0.3} & \textbf{88.0}\,\std{1.7} & \ul{46.1}\,\std{1.4} & \textbf{58.8}\,\std{2.9} & \textbf{61.5}\,\std{1.2} & \textbf{75.0} & \textbf{55.0} & \textbf{85.0} & \textbf{65.0} & \textbf{70.0} \\
\bottomrule
\end{tabular}
\vspace{-0.2cm}
\end{table*}

\subsection{Experimental Setup}
\label{sec:setup}

\subsubsection{Benchmarks}
We evaluate \ourmodel in simulation and the real world with diverse training-time camera viewpoints.
In simulation, we use 24 RoboCasa tasks~\cite{robocasa2024} with 50 human demonstrations per task.
Based on the RoboCasa atomic-skill taxonomy, we group them into four categories: pick-and-place (8 tasks), drawer/door opening and closing (6), knob/lever turning (5), and button pressing/insertion (5).
The external cameras are randomized per demonstration and fixed to reference viewpoints at evaluation, following RoboCasa's default setup (\Cref{fig:setup}, left).
We report mean and standard deviation over three seeds, with 50 episodes per task per seed.
For real-world experiments, we use a Franka Research 3 with a wrist camera (RealSense D405) and an external camera (RealSense D435i).
We collect 320 demonstrations across four tasks and eight external camera viewpoints, with 10 demonstrations per task per viewpoint (\Cref{fig:setup}, right).
We evaluate each method over 20 trials per task at one training viewpoint and report success rates.

\subsubsection{Baselines}
We compare \ourmodel against RGB-only $\pi_{0.5}$~\cite{black2025pi05} and three groups of baselines incorporating viewpoint or 3D information.
\textit{(i) Camera-geometry-aware VLAs} include KYC~\cite{jiang2026knowyourcamera}, which provides per-pixel Pl\"ucker rays from camera intrinsics and extrinsics, and OC-VLA~\cite{ocvla}, which predicts actions in the camera frame.
\textit{(ii) Scene-geometry-augmented VLAs}, including GeoVLA~\cite{geovla} and PointVLA~\cite{pointvla}, process point clouds with dedicated 3D modules.
\textit{(iii) Point cloud policies} are represented by FP3~\cite{fp3}, which uses robot-frame point clouds without a pretrained VLM backbone.
As official implementations are unavailable for groups \textit{(i)} and \textit{(ii)}, we reimplement them on the $\pi_{0.5}$ backbone for controlled comparison.
For real-world experiments, we select OC-VLA and PointVLA as the strongest variants in their respective groups in simulation, alongside RGB-only $\pi_{0.5}$ and FP3.

\subsubsection{Implementation Details}
For the main experiments, we initialize $\pi_{0.5}$-based methods from pretrained $\pi_{0.5}$ (base) and FP3 from its DROID-pretrained checkpoint~\cite{khazatsky2024droid}.
We fine-tune them for 20k steps on RoboCasa and 50k steps on real-world data with a batch size of 64.
The pointmap encoder is initialized from the pretrained RGB encoder~\cite{zhai2023siglip}, and both are updated during training.
The $\pi_{0.5}$-based methods use AdamW~\cite{loshchilov2017decoupled} ($\beta=(0.9,0.95)$, weight decay 0.01) with cosine decay and learning rates of $10^{-4}$ in simulation and $2.5\times10^{-5}$ in the real world.
To isolate design choices from robot pretraining, ablations in \Cref{sec:ablation} and \Cref{sec:origin_ablation} use a $\pi_{0.5}$-style architecture~\cite{black2025pi05} initialized from base PaliGemma~\cite{beyer2024paligemma}, with all variants trained for 30k steps.

Following OpenPI~\cite{black2025pi05}, $\pi_{0.5}$-based methods condition on $o_t$, predict 50 actions, and execute 25 before replanning.
Following its official setup, FP3 conditions on $(o_{t-1},o_t)$ with 4000 points per camera, predicts 16 actions, and executes 8.
For all real-world depth-based inputs, we use Fast-FoundationStereo~\cite{wen2026fastfoundationstereo} to reduce sensor noise.
We discard depth beyond 10\,m in simulation and 3\,m for real-world external cameras; missing or discarded pixels are represented by a learned three-dimensional mask token in the pointmap.

\subsection{Comparison with 3D-augmented VLAs}
\label{sec:main-exp}

\subsubsection{Simulation}
\label{subsec:main-exp-sim}

Among methods sharing the $\pi_{0.5}$ backbone, the strongest camera-aware variant (OC-VLA~\cite{ocvla}, $57.4$) and 3D-augmented variant (PointVLA~\cite{pointvla}, $58.0$) both remain below \ourmodel ($61.5$).
Notably, both PointVLA and FP3~\cite{fp3} process geometry as point clouds rather than dense, RGB-aligned pointmaps.
Their lower performance suggests that preserving dense spatial correspondence between RGB and geometry may be beneficial for manipulation.
These methods differ in both geometric inputs and integration mechanisms, so \Cref{sec:ablation} isolates these design choices through controlled comparisons.
FP3, the only baseline not built on $\pi_{0.5}$, further lacks large-scale vision-language pretraining and uses a sparse point cloud input (4000 points per view), compared with our $128\times128$ pointmap (16,384 points per view).

\subsubsection{Real-world}
\label{subsec:main-exp-real}

\Cref{tab:main_robocasa} shows that adding robot-centric pointmaps raises the average success rate of $\pi_{0.5}$ from $37.5$ to $70.0$, achieving the highest success rate on all four tasks.
PointVLA~\cite{pointvla}, the strongest baseline, reaches only $46.3$ on average.
The benefit of explicit 3D geometry is particularly evident on peg-in-hole, where PointVLA, FP3~\cite{fp3}, and \ourmodel achieve $35.0\%$, $55.0\%$, and $65.0\%$, respectively, compared with $20.0\%$ for both RGB-only $\pi_{0.5}$ and OC-VLA~\cite{ocvla}.
While FP3 performs strongly on drawer opening ($80.0\%$) and peg-in-hole ($55.0\%$), it drops to $25.0\%$ on pick-and-place and $15.0\%$ on stacking.
Our $640\times480$ pointmaps retain 307,200 points per view, over $75\times$ more than the 4000-point clouds used by PointVLA and FP3, while preserving pixel-wise correspondence with RGB.
We further observe that FP3 failures often involve semantic object selection, such as choosing the wrong object or target.
These results suggest that combining explicit geometry for precise manipulation with pretrained vision-language features for semantic reasoning is important, with \ourmodel consistently improving performance across all four tasks.

\begin{figure*}[t]
\centering
\includegraphics[width=0.95\linewidth]{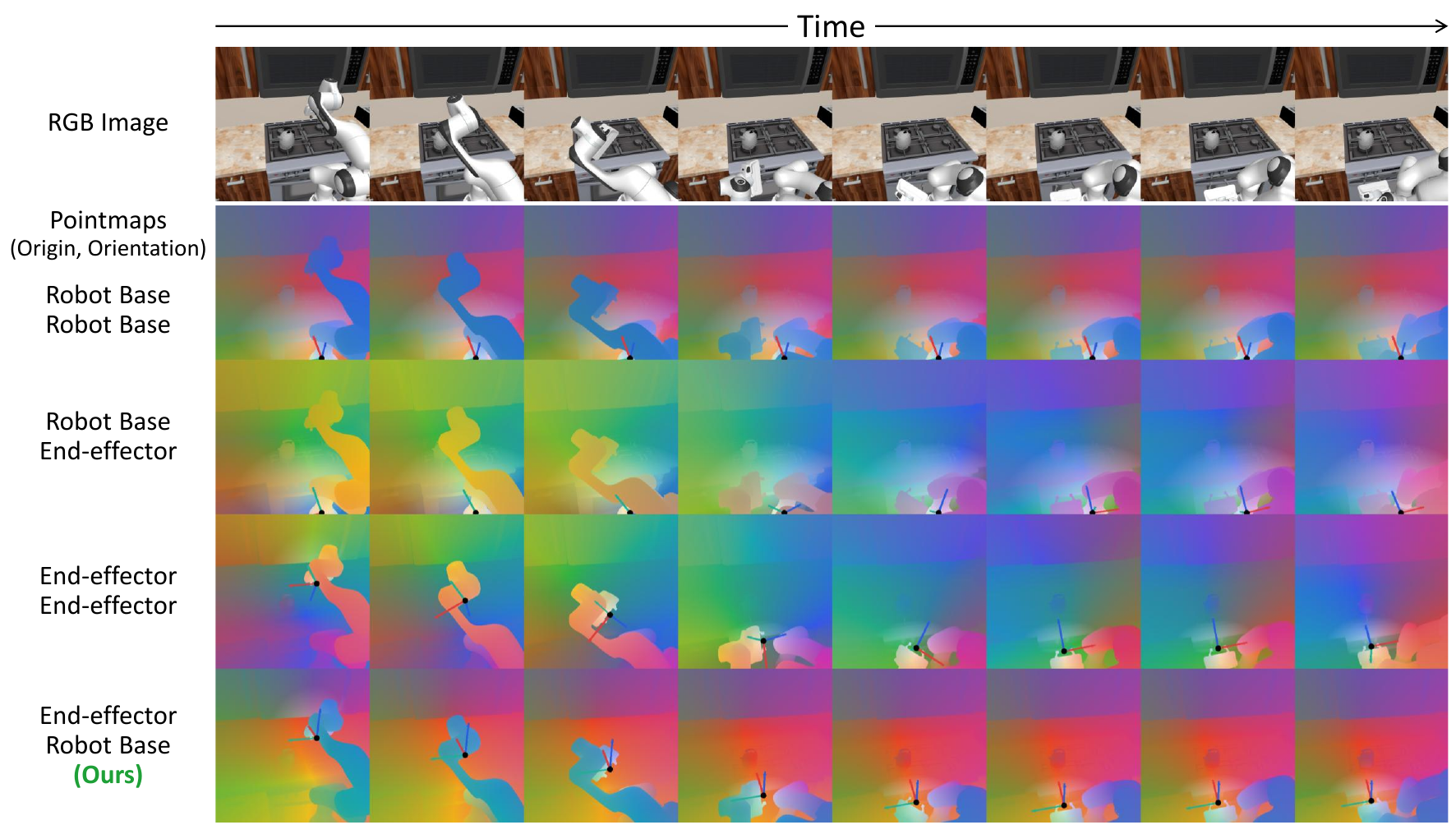}
\caption{\textbf{Qualitative comparison of pointmap coordinate frames along a RoboCasa rollout.}
Each row specifies the origin followed by the axis orientation.
An end-effector origin expresses objects relative to the gripper, providing a common reference for similar interactions across workspace locations.
End-effector-aligned axes introduce wrist-induced coordinate changes (middle two pointmap rows), whereas our design preserves gripper-relative positions without these changes by retaining robot-base-aligned axes (bottom row).}
\label{fig:robotframe-pointmap}
\vspace{-0.4cm}
\end{figure*}

\subsection{Ablation Studies}
\label{sec:ablation}

To examine the factors underlying these gains, we conduct controlled ablations on 3D representations, pointmap--RGB fusion, and training-time viewpoint variation.
To study these factors without inheriting task and viewpoint priors from a particular robot-pretraining dataset, all experiments in~\Cref{sec:ablation,sec:origin_ablation} initialize $\pi_{0.5}$ from pretrained VLM weights~\cite{beyer2024paligemma} without robot-action pretraining.
For the representation comparison, we fix the fusion method to addition.
For the fusion comparison, we fix the representation to pointmaps.
These two comparisons use the robot-base frame for representations requiring a 3D coordinate frame, while coordinate-frame design is studied separately in \Cref{sec:origin_ablation}.

\subsubsection{How should 3D geometry be represented?}
\label{subsubsec:modality}

We first compare depth, camera, point clouds, and pointmaps as geometric inputs to the VLA in \Cref{tab:ablation_modality_fusion}.
Depth provides per-pixel surface distance, while Pl\"ucker rays~\cite{jiang2026knowyourcamera} encode camera geometry from intrinsics and extrinsics.
Together, these signals contain the information required to recover 3D scene coordinates in the robot frame.
Depth and Pl\"ucker rays individually improve over RGB alone ($29.6$ and $29.3$ vs.\ $25.6$), while combining them further increases success to $31.7$.
Explicitly converting the same information into robot-frame pointmaps further improves success to $34.4$, suggesting a benefit from providing robot-aligned 3D coordinates rather than leaving their geometric relationship implicit.

We also compare against a point-cloud representation.
We encode robot-frame points with a pretrained point-cloud encoder~\cite{wu2025sonata}, align the resulting features with RGB patches following~\cite{fan2026any3d}, and add them to the corresponding RGB tokens.
This variant reaches $31.4$, below the pointmap representation.
Despite both providing robot-frame 3D geometry, the pointmap retains dense pixel-wise correspondence with RGB, which may contribute to its higher performance.
Because the two variants also differ in their geometric encoders, this comparison does not isolate the effect of density alone.
Overall, these results motivate our use of dense, precomputed robot-centric pointmaps.

\begin{table}[t]
\centering
\caption{
Controlled Ablation of 3D Modality and Fusion Method.
}
\label{tab:ablation_modality_fusion}
\vspace{-0.15cm}
\begin{tabular}{lccc}
\toprule
3D Modality & Fusion Method & \begin{tabular}[c]{@{}c@{}}SR (\%)\\(24 Tasks)\end{tabular} & \begin{tabular}[c]{@{}c@{}}$\Delta$\\ vs. Ref.\end{tabular} \\
\drule
\g RGB-only & --- & 25.6\,\std{1.3} & $-$8.8 \\
\g Depth & Addition & 29.6\,\std{1.1} & $-$4.8 \\
\g Pl\"ucker & Addition & 29.3\,\std{0.9} & $-$5.1 \\
\g Depth \& Pl\"ucker & Addition & 31.7\,\std{0.3} & $-$2.7 \\
\g Point Cloud & Addition & 31.4\,\std{1.5} & $-$3.0 \\
\midrule
Pointmap & \g Token Concat. & 30.3\,\std{0.6} & $-$4.1 \\
Pointmap & \g Channel Concat. & 31.6\,\std{1.4} & $-$2.8 \\
\midrule
Pointmap & Addition & \textbf{34.4}\,\std{1.3} & --- \\
\bottomrule
\end{tabular}
\end{table}

\subsubsection{How should pointmaps be fused with RGB?}
\label{subsubsec:fusion_ablation}

We next compare channel-wise concatenation, token-wise concatenation, and addition for fusing pointmaps with RGB in \Cref{tab:ablation_modality_fusion}.
Among the three strategies, addition performs best, reaching $34.4$ compared with $30.3$ for token-wise concatenation and $31.6$ for channel-wise concatenation.
Unlike channel-wise concatenation, addition allows RGB and pointmaps to be encoded separately while preserving one-to-one spatial correspondence between their visual tokens.
It also avoids increasing the visual token sequence length, as in token-wise concatenation.
These results suggest that separately encoding the two modalities and combining spatially aligned features is an effective way to incorporate pointmaps into the VLA.

\begin{figure}[t]
\centering
\vspace{0.3cm}
\includegraphics[width=\linewidth]{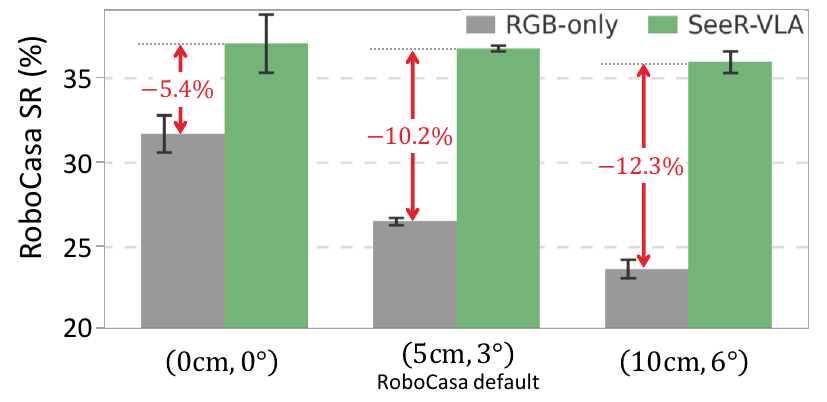}
\vspace{-0.3cm}
\caption{\textbf{Increasing gains with training-time camera variation.}
External-camera poses are randomized per demonstration with independent zero-mean Gaussian noise using per-axis translation/rotation standard deviations of $(0\,\mathrm{cm},0^\circ)$, $(5\,\mathrm{cm},3^\circ)$, or $(10\,\mathrm{cm},6^\circ)$, while the wrist camera remains unchanged.
As variation increases, \ourmodel maintains success while RGB-only performance declines, widening the gap.}
\label{fig:cam_variation}
\end{figure}

\subsubsection{How does training-time viewpoint variation affect performance?}
\label{subsubsec:camera_variation}

We evaluate whether robot-centric pointmaps provide larger gains over RGB-only input as camera viewpoints in the training data become more diverse.
For each training demonstration, we perturb the external-camera poses with independent zero-mean Gaussian noise, using per-axis translation/rotation standard deviations of $(0\,\mathrm{cm},0^\circ)$, $(5\,\mathrm{cm},3^\circ)$, or $(10\,\mathrm{cm},6^\circ)$.
The wrist camera and other training settings remain unchanged.
As shown in \Cref{fig:cam_variation}, the success-rate gap between \ourmodel and RGB-only increases from $5.4$ to $10.2$ and $12.3$ percentage points as viewpoint variation increases.
RGB-only performance declines, whereas \ourmodel remains comparatively stable.
These results suggest that providing 3D geometry in robot coordinates is particularly useful when learning from demonstrations collected across diverse camera viewpoints, 
making it well suited to training on large-scale robot datasets.

\subsection{Which Coordinate Frame Should Pointmaps Use?}
\label{sec:origin_ablation}

A pointmap coordinate frame is determined by its origin and axis orientation.
We vary each independently between the robot base and end effector while fixing the pointmap representation and additive RGB fusion.

\subsubsection{Qualitative Analysis}

\Cref{fig:robotframe-pointmap} illustrates how the origin and orientation affect pointmaps along a rollout.
The first and last pointmap rows isolate the effect of changing the origin while retaining robot-base-aligned axes.
With a robot-base origin, point coordinates describe absolute workspace locations, so the same gripper--object configuration can be represented differently when it occurs at different locations.
Centering the pointmap at the end effector instead keeps the gripper at the origin and represents surrounding points by their displacement from it.
Thus, similar interactions share similar geometric coordinates even across different workspace locations.

The middle two rows illustrate the effect of rotating the axes with the end effector.
As the wrist rotates, end-effector-aligned axes rotate as well, causing coordinates to change even for static scene points.
Moreover, the meaning of each coordinate axis changes with wrist orientation; for example, a positive $z$ coordinate no longer consistently denotes the same robot-frame direction.
Using robot-base-aligned axes avoids these wrist-induced changes while retaining the gripper-relative positions provided by end-effector centering.
Our coordinate frame therefore combines a moving origin for interaction-relative geometry with fixed axes for consistent spatial directions.

\begin{table}[t]
\centering
\vspace{-0.1cm}
\caption{
Controlled Ablation of the Pointmap's Coordinate Frames.
}
\label{tab:ablation_coord_frame}
\vspace{-0.15cm}
\begin{tabular}{cccc}
\toprule
\multicolumn{2}{c}{Coordinate Frame} & \multirow{2}{*}[-0.1cm]{\begin{tabular}[c]{@{}c@{}}SR (\%)\\(24 Tasks)\end{tabular}} & \multirow{2}{*}[-0.1cm]{\begin{tabular}[c]{@{}c@{}}$\Delta$\\ vs. Ours\end{tabular}} \\
\cmidrule(lr){1-2}
Origin & Orientation &  &  \\
\drule
\g Robot Base & Robot Base & 34.4\,\std{1.3} & $-$2.4 \\
\g Robot Base & \g End-effector & 33.4\,\std{1.4} & $-$3.4 \\
End-effector & \g End-effector & \ul{34.9}\,\std{1.1} & $-$1.9 \\
\midrule
End-effector & Robot Base & \textbf{36.8}\,\std{0.2} & --- \\
\bottomrule
\end{tabular}
\vspace{-0.1cm}
\end{table}

\subsubsection{Quantitative Comparison}

\Cref{tab:ablation_coord_frame} confirms this qualitative intuition.
Moving the origin to the end effector improves success under both orientations, from $34.4\%$ to $36.8\%$ with robot-base axes and from $33.4\%$ to $34.9\%$ with end-effector axes.
In contrast, aligning the axes with the end effector reduces success under both origins.
The combination of an end-effector origin and robot-base axes performs best at $36.8\%$, supporting the use of gripper-relative coordinates with stable robot-base-aligned directions.

\subsection{Analysis}
\label{sec:analysis-sec}

\begin{figure}[t]
  \centering
\includegraphics[width=\linewidth]{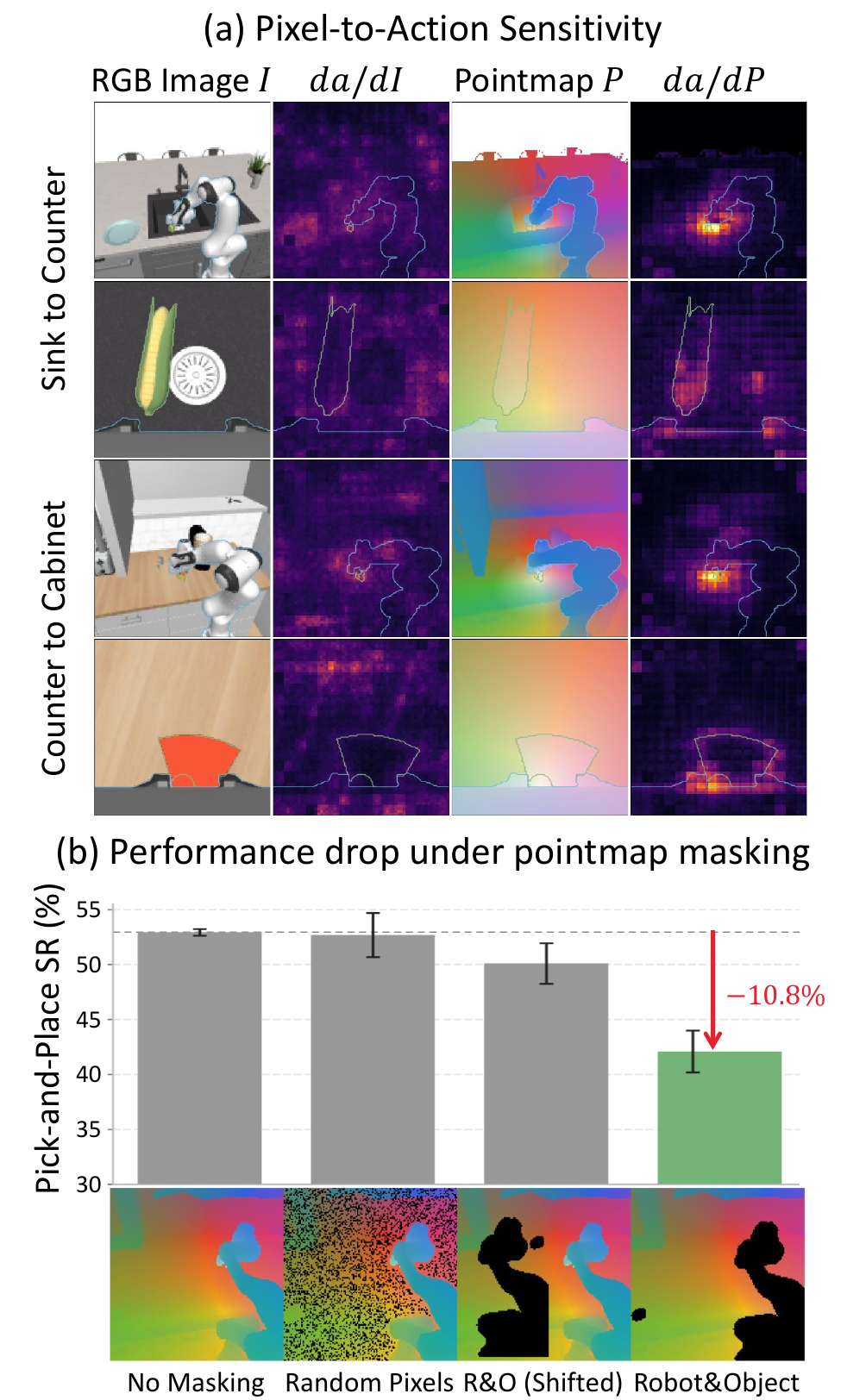}
  \caption{\textbf{Policy reliance on task-relevant pointmap geometry.}
    Masking target-object and robot regions in pointmaps lowers success on eight pick-and-place tasks, unlike equal-sized shifted or random masks.
    Action gradients with respect to pointmap coordinates, $|\partial a/\partial P(u,v)|$, highlight target pixels where small coordinate changes most affect predicted actions.}
  \label{fig:masking}
\end{figure}

Having examined the pointmap design, we now investigate whether the policy exploits object and robot coordinates in the pointmap to predict actions.
We first examine whether \ourmodel uses pointmap geometry and which regions influence its actions.
For observations from evaluation episodes, we compute gradients of the predicted end-effector motion $a$ over the next 25 steps with respect to pointmap coordinates and RGB values.
The resulting magnitudes, $|\partial a/\partial P(u,v)|$ and $|\partial a/\partial I(u,v)|$, indicate where small input changes most affect the predicted motion.
As shown in~\Cref{fig:masking}~(a), pointmap sensitivity concentrates on the target object outlined in green, whereas RGB sensitivity is more broadly distributed.
This localization suggests that \ourmodel uses the target's 3D coordinates for action prediction.

We next test whether removing this geometry impairs task execution.
We replace robot-frame pointmap coordinates covering the target object and robot with a mask token while leaving RGB unchanged, and evaluate on eight RoboCasa pick-and-place tasks.
This targeted masking reduces success from $52.9\%$ to $42.1\%$ (\Cref{fig:masking}(b)).
To determine whether this drop reflects the removal of task-relevant geometry rather than a generic masking effect, we evaluate two additional masking strategies.
Shifted masking moves the same mask to a random, task-irrelevant location, while random masking removes the same number of scattered pixels.
These strategies retain success rates of $51.5\%$ and $52.5\%$, respectively, despite masking as many pixels as the target-object and robot mask.
Together, the gradient maps and masking results show that \ourmodel relies on target-object and robot coordinates for action prediction.

\section{Conclusion}
\label{sec:conclusion}

We study how 3D geometry should be represented and integrated into VLAs trained on robot data collected from diverse camera viewpoints.
Our results show that explicitly expressing geometry as dense, robot-centric pointmaps is more effective than leaving geometric relationships implicit or processing geometry as separate point clouds.
Controlled ablations demonstrate the effectiveness of spatially aligned fusion of separately encoded RGB and pointmap features, the benefits of end-effector-centered coordinates with robot-base-aligned axes, and increasing pointmap gains as training viewpoints diversify.
This design improves pretrained VLAs in both simulation and real-world experiments, with analysis of the final policy demonstrating reliance on task-relevant pointmap geometry.
While our method currently requires metric depth and camera calibration, future work could reduce these requirements by leveraging monocular geometry models~\cite{vggt, moge,moge2} to estimate metric depth and intrinsics alongside camera-to-robot pose estimation.
Overall, our results highlight that both the representation and coordinate frame of 3D observations matter, and that robot-centric pointmaps provide a simple way to align visual geometry with the space in which the robot acts.

\bibliographystyle{IEEEtran}
\bibliography{root}

\end{document}